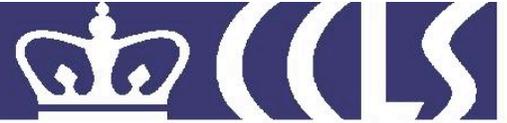

TECHNICAL REPORT

**No.** CCLS-13-02
**Title**: LDC Arabic Treebanks and Associated Corpora:
Data Divisions Manual
**Authors**: Mona Diab, Nizar Habash, Owen Rambow and
Ryan Roth

# LDC Arabic Treebanks and Associated Corpora: Data Divisions Manual

Mona Diab and Nizar Habash and Owen Rambow and Ryan Roth
Center for Computational Learning Systems
Columbia University
`{diab,habash,rambow,ryanr}@ccls.columbia.edu`

Version: 1.0

## 1    Introduction

The Linguistic Data Consortium (LDC) has developed hundreds of data corpora for natural language processing (NLP) research. Among these are a number of annotated treebank corpora for Arabic. Typically, these corpora consist of a single collection of annotated documents. NLP research, however, usually requires multiple data sets for the purposes of training models, developing techniques, and final evaluation. Therefore it becomes necessary to divide the corpora used into the required data sets (divisions).

Unfortunately, there is no universally accepted convention or standard for dividing bulk corpora. This caused different research groups to either define their own divisions (which makes comparison to similar research results difficult) or adopt existing published divisions (which do not adapt as new corpora versions are released). When a new treebank is released, a new division needs to be developed, which may or may not be consistent with the other treebank divisions.

This document details a set of rules that have been defined to enable consistent divisions for old and new Arabic treebanks (ATB) and related corpora. These rules have been applied to the currently available LDC Modern Standard Arabic Treebanks (ATB1 - ATB12) [7, 8], the Egyptian Arabic Treebanks (ARZ1 - ARZ8) [9] and the spoken Levantine ATB [6], and the exact divisions are listed in tables.

## 2    Arabic Treebanks and Related Corpora

The LDC has released a number of annotated treebanks for Modern Standard Arabic; this paper only examines the first 12; note that, historically, ATB1, ATB2 and ATB3 have been the most frequently utilized and updated. In addition to the MSA ATBs, this paper examines the first 8 Egyptian Treebanks (ARZ1 - ARZ8), and an older spoken Levantine treebank.

Each treebank has gone through one or more versions; new versions typically only alter the annotations in the treebank, and do not usually alter the raw text or documents used. This means that a division defined for a treebank can usually be safely applied to future versions of the same treebank. However, if a new version adds or removes documents then the data division needs to be redefined.

The treebanks examined, their versions and their LDC catalog numbers are listed in Table 1.

## 3    Division Rules

In defining a set of data division rules, there are several goals. First, at a minimum there needs to be a defined training set (TRAIN), a defined development set (DEV) and a defined testing set (TEST).

Second, the DEV and TEST sections should be taken from well-separated regions of the corpora, in order to ensure that similar results derived from each are not due to mere proximity of subject matter.

Third, divisions should be applied across document boundaries, not sentence or word boundaries, to prevent the need to break apart documents.

Fourth, the divisions should not utilize any random processes in selection; randomly selecting documents for each division may or may not improve model performance, but will make describing the data divisions in publications more difficult.

Fifth, the defined TEST divisions should be at least as large as the largest (commonly-used) TEST divisions previously defined for that treebank, and include the same files. By requiring this, comparisons to previous results using those previous TEST divisions are still possible, simply by excluding any extra documents in the current TEST division.

Finally, each TRAIN division should hold roughly 80% of the corpora by word volume, with the remainder split across the other divisions.

These goals are accomplished by the following procedure:

| Treebank Label | Version | LDC Catalog Number |
|---|---|---|
| **MSA Treebanks** | | |
| ATB1 | 4.1 | LDC2010T13 |
| ATB2 | 3.1 | LDC2011T09 |
| ATB3 | 3.2 | LDC2010T08 |
| ATB4 | 1.0 | LDC2005T30 |
| ATB5 | 1.0 | LDC2009E72 |
| ATB6 | 1.0 | LDC2009E108 |
| ATB7 | 1.0 | LDC2009E114 |
| ATB8 | 1.0 | LDC2010E11 |
| ATB9 | 1.0 | LDC2010E19 |
| ATB10 | 1.0 | LDC2010E22 |
| ATB11 | 1.0 | LDC2011E16 |
| ATB12 | 1.0 | LDC2011E17 |
| **Egyptian Treebanks** | | |
| ARZ1 | 2.0 | LDC2012E93 |
| ARZ2 | 2.0 | LDC2012E98 |
| ARZ3 | 2.0 | LDC2012E89 |
| ARZ4 | 2.0 | LDC2012E99 |
| ARZ5 | 2.0 | LDC2012E107 |
| ARZ6 | 2.0 | LDC2012E125 |
| ARZ7 | 2.0 | LDC2013E12 |
| ARZ8 | 2.0 | LDC2013E21 |
| **Levantine Treebanks** | | |
| Spoken Levantine | 2.0 | LDC2005E78 |

Table 1: Considered LDC Arabic Treebanks and Associated Corpora

1. The documents in the treebank are sorted by document name

2. Documents from start of the document list (that is, those alphabetically first) are set into the DEV division until the total word count in DEV exceeds 10% of the total corpora word count

3. Documents from the end of the document list are set into the TEST division until the total word count in TEST exceeds 10% of the total corpora word count

4. The remaining documents (roughly 80% of the total word count) are set into the TRAIN division

If a particular project requires additional divisions beyond TRAIN, DEV and TEST, the project should take the needed documents from the TRAIN division, to ensure that the DEV and TEST divisions remain constant and stable.

# 4    Arabic Treebank Divisons

Tables 2 - 8 list the exact document divisions produced for each treebank when the above division rules are applied. The Document Range column lists the beginning and ending document for that data subset. The number of documents and number of raw words contained in each data subset are also listed.

Note that ATB6 is a special case; this MSA treebank contains text taken from newswire, newsgroup and weblog genre sources, which often need to be isolated from each other. Therefore each of these genres is treated as an independent sub-treebank, with its own DEV, TRAIN, and TEST divisions.

| Division | # Docs | # Words | Document Range |
|----------|--------|---------|----------------|
| **ATB1 DEV** | 71 | 14635 | 20000715_AFP_ARB.0001 - 20000715_AFP_ARB.0073 |
| **ATB1 TRAIN** | 568 | 116198 | 20000715_AFP_ARB.0074 - 20001115_AFP_ARB.0128 |
| **ATB1 TEST** | 95 | 14553 | 20001115_AFP_ARB.0129 - 20001115_AFP_ARB.0236 |
| **ATB2 DEV** | 50 | 14468 | UMAAH_UM.ARB_20010721-e.0018 - UMAAH_UM.ARB_20020217-a.0019 |
| **ATB2 TRAIN** | 400 | 115147 | UMAAH_UM.ARB_20020224-a.0005 - UMAAH_UM.ARB_backissue_34-a.0013 |
| **ATB2 TEST** | 51 | 14584 | UMAAH_UM.ARB_backissue_34-a.0014 - UMAAH_UM.ARB_backissue_40-e.0025 |
| **ATB3 DEV** | 58 | 34033 | ANN20020115.0001 - ANN20020115.0110 |
| **ATB3 TRAIN** | 480 | 271646 | ANN20020215.0001- ANN20021115.0033 |
| **ATB3 TEST** | 61 | 34031 | ANN20021115.0034 - ANN20021215.0045 |

Table 2: ATB 1,2,3 Data Divisions. These are the most commonly used ATBs, and are often combined. The current versions are ATB1 V4.1, ATB2 V3.1, and ATB3 V3.2.

| Division | # Docs | # Words | Document Range |
|----------|--------|---------|----------------|
| **ATB4 DEV** | 41 | 16344 | ASB20040928.0001 - ASB20041003.0004 |
| **ATB4 TRAIN** | 315 | 129310 | ASB20041003.0005 - ASB20041101.0001 |
| **ATB4 TEST** | 41 | 16261 | ASB20041101.0002 - ASB20041104.0019 |
| **ATB5 DEV** | 3 | 12442 | ALHURRA_NEWS10_ARB_20051127_100101 - ALHURRA_NEWS13_ARB_20051121_130100 |
| **ATB5 TRAIN** | 24 | 77257 | ALHURRA_NEWS13_ARB_20051123_130100.qrtr - DUBAI_DUBAINEWS_ARB_20050223_113000 |
| **ATB5 TEST** | 4 | 10927 | DUBAI_DUBAINEWS_ARB_20050228_113000 - DUBAI_DUBAINEWS_ARB_20051219_112801 |

Table 3: ATB4 and ATB5 Data Divisions. Both of these treebanks are currently version 1.0.

| Division | # Docs | # Words | Document Range |
|---|---|---|---|
| **NEWSWIRE** | | | |
| **ATB6 NW DEV** | 13 | 2910 | AAW_ARB_20080502.0027-S1 - AFP_ARB_20080510.0077-S1 |
| **ATB6 NW TRAIN** | 96 | 21738 | AFP_ARB_20080512.0051-S1 - QDS_ARB_20080517.0021-S1 |
| **ATB6 NW TEST** | 13 | 2836 | QDS_ARB_20080517.0046-S1 - XIN_ARB_20080522.0166-S1 |
| **NEWSGROUP** | | | |
| **ATB6 NG DEV** | 4 | 873 | arb-NG-2-76511-10666609-S1 - arb-NG-2-76513-10114621-S1 |
| **ATB6 NG TRAIN** | 27 | 6124 | arb-NG-2-76513-10667328-S1 - arb-NG-31-114370-10352403-S2 |
| **ATB6 NG TEST** | 4 | 1018 | arb-NG-95-174352-11045427-S1 - arb-NG-95-174357-11048231-S2 |
| **WEBLOG** | | | |
| **ATB6 WL DEV** | 9 | 1948 | arb-WL-1-152170-10153737-S2 - arb-WL-1-152350-10158443-S1 |
| **ATB6 WL TRAIN** | 68 | 15430 | arb-WL-1-152375-10682332-S1 - arb-WL-7-88414-10669660-S2 |
| **ATB6 WL TEST** | 9 | 2064 | arb-WL-7-88466-10669891-S1 - arb-WL-7-89199-10673129-S1 |

Table 4: ATB6 V1.0 Data Divisions. This treebank has multiple genres, which typically need to be isolated from each other. Therefore, each genre (newswire, newsgroup, weblog) is treated as a separate sub-treebank here.

| Division | # Docs | # Words | Document Range |
|---|---|---|---|
| **ATB7 DEV** | 1 | 4052 | ALFAYHA_NEWS_ARB_20080401_210000 |
| **ATB7 TRAIN** | 10 | 25723 | ALHURRA_THEGLOBAL_ARB_20080118_210000 - SAWA_SAWANEWS_ARB_20080208_181500 |
| **ATB7 TEST** | 4 | 4589 | SAWA_SAWANEWS_ARB_20080304_221500 - SAWA_SAWANEWS_ARB_20080307_221500 |
| **ATB8 DEV** | 2 | 8109 | ABUDHABI_ABUDHNEWS2_ARB_20070228_000000.qrtr - ABUDHABI_ABUDHNEWS_ARB_20070110_115800.qrtr |
| **ATB8TRAIN** | 12 | 57750 | ABUDHABI_ABUDHNEWS_ARB_20070117_115800.qrtr - ALHURRA_THEWORLDTODAY_ARB_20080208_170000 |
| **ATB8 TEST** | 4 | 9275 | ALURDUNYA_URDUNYANEWS_ARB_20070312_000000.qrtr - ARABIYA_ALARABIYANEWS2_ARB_20070312_000000.qrtr |
| **ATB9 DEV** | 3 | 10966 | ARABIYA_ALARABIYANEWS2_ARB_20070316_000000.qrtr - ARABIYA_ALARABIYANEWS2_ARB_20080409_200000 |
| **ATB9 TRAIN** | 13 | 53955 | ARABIYA_LATEHRNEWS_ARB_20070222_000000.qrtr - SYRIANTV_NEWS25_ARB_20070122_162800.qrtr |
| **ATB9 TEST** | 2 | 12593 | SYRIANTV_NEWS25_ARB_20070129_162800.qrtr - SYRIANTV_NEWS25_ARB_20070201_162801.qrtr |
| **ATB10 DEV** | 1 | 5972 | ALHURRA_THEGLOBAL_ARB_20080205_210000 |
| **ATB10 TRAIN** | 7 | 30961 | ALJZ_TODHARV_ARB_20070107_145800.qrtr - SAUDITV_SAUDINEWS2_ARB_20080326_190000 |
| **ATB10 TEST** | 1 | 4462 | SAUDITV_SAUDINEWS2_ARB_20080402_200000 |
| **ATB11 DEV** | 11 | 4412 | ABDULEMAM_20041226.1648 - DIGRESSING_20041107.0106 |
| **ATB11 TRAIN** | 83 | 30195 | DIGRESSING_20041109.0437 - TAREEKALSHAAB_20041114.1958 |
| **ATB11 TEST** | 11 | 4080 | TAREEKALSHAAB_20041114.1959 - ZAYEDALSAIDI_20050221.1414 |
| **ATB12 DEV** | 3 | 15470 | ABUDHABI_ABUDHNEWS_ARB_20070111_115801.qrtr - ALAM_NEWSRPT_ARB_20070102_015800.qrtr |
| **ATB12 TRAIN** | 25 | 88367 | ALAM_NEWSRPT_ARB_20070111_015800.qrtr - SCOLA_JORDNNSCO_ARB_20070308_095800.qrtr |
| **ATB12 TEST** | 3 | 12565 | SCOLA_SAUDNNSCO_ARB_20070222_215800.qrtr - SYRIANTV_NEWS25_ARB_20070208_162800.qrtr |

Table 5: ATB7 - ATB12 Data Divisions. All of these treebanks are currently version 1.0.

| Division | # Docs | # Words | Document Range |
|----------|--------|---------|----------------|
| **ARZ1 DEV** | 4 | 3842 | bolt-arz-NG-169-181081-14577.arz.su - <br> bolt-arz-NG-169-181081-19026.arz.su |
| **ARZ1 TRAIN** | 46 | 28837 | bolt-arz-NG-169-181081-21390.arz.su - <br> bolt-arz-NG-169-181090-38942.arz.su |
| **ARZ1 TEST** | 8 | 4078 | bolt-arz-NG-169-181090-38993.arz.su - <br> bolt-arz-NG-169-181090-40037.arz.su |
| **ARZ2 DEV** | 4 | 3280 | bolt-arz-NG-169-181081-16222.arz.su - <br> bolt-arz-NG-169-181081-68225.arz.su |
| **ARZ2 TRAIN** | 31 | 22201 | bolt-arz-NG-169-181081-68287.arz.su - <br> bolt-arz-NG-169-181090-39607.arz.su |
| **ARZ2 TEST** | 3 | 3732 | bolt-arz-NG-169-181090-39695.arz.su - <br> bolt-arz-NG-169-181090-40322.arz.su |
| **ARZ3 DEV** | 8 | 3682 | bolt-arz-DF-175-182187-572764.arz.su - <br> bolt-arz-DF-175-182187-577973.arz.su |
| **ARZ3 TRAIN** | 43 | 24067 | bolt-arz-DF-175-182187-578399.arz.su - <br> bolt-arz-NG-169-181090-40341.arz.su |
| **ARZ3 TEST** | 5 | 3762 | bolt-arz-NG-169-181090-40504.arz.su - <br> bolt-arz-NG-169-181092-26920.arz.su |
| **ARZ4 DEV** | 5 | 6321 | bolt-arz-DF-175-182187-575959.arz.su - <br> bolt-arz-DF-175-182187-581488.arz.su |
| **ARZ4 TRAIN** | 64 | 30285 | bolt-arz-DF-175-182187-581658.arz.su - <br> bolt-arz-DF-175-182192-10963633.arz.su |
| **ARZ4 TEST** | 7 | 4855 | bolt-arz-DF-175-182258-1245345.arz.su - <br> bolt-arz-NG-169-181090-40249.arz.su |
| **ARZ5 DEV** | 9 | 4021 | bolt-arz-DF-169-181089-15751715.arz.su - <br> bolt-arz-DF-169-181091-8751442.arz.su |
| **ARZ5 TRAIN** | 73 | 28361 | bolt-arz-DF-175-182185-10963619.arz.su - <br> bolt-arz-DF-204-185979-1392879.arz.su |
| **ARZ5 TEST** | 13 | 4037 | bolt-arz-DF-204-185979-1393182.arz.su - <br> bolt-arz-NG-169-181081-72955.arz.su |
| **ARZ6 DEV** | 19 | 10542 | bolt-arz-DF-169-181090-8816189.arz.su - <br> bolt-arz-DF-175-182187-572570.arz.su |
| **ARZ6 TRAIN** | 164 | 77382 | bolt-arz-DF-175-182187-572693.arz.su - <br> bolt-arz-DF-207-186125-504972.arz.su |
| **ARZ6 TEST** | 16 | 10997 | bolt-arz-DF-207-186125-506363.arz.su - <br> bolt-arz-NG-169-181090-40862.arz.su |

Table 6: Egyptian Arabic Treebank Data Divisions (ARZ 1-6). All above treebanks are currently version 2.0.

| Division | # Docs | # Words | Document Range |
|---|---|---|---|
| **ARZ7 DEV** | 4 | 6462 | bolt-arz-DF-175-182187-572367.arz.su - bolt-arz-DF-175-182187-573032.arz.su |
| **ARZ7 TRAIN** | 76 | 47175 | bolt-arz-DF-175-182187-575718.arz.su - bolt-arz-DF-210-186177-3027514.arz.su |
| **ARZ7 TEST** | 13 | 7475 | bolt-arz-DF-210-186206-3701939.arz.su - bolt-arz-DF-217-194296-6676523.arz.su |
| **ARZ8 DEV** | 15 | 6895 | bolt-arz-DF-175-182187-572370.arz.su - bolt-arz-DF-175-182188-1048352.arz.su |
| **ARZ8 TRAIN** | 91 | 51085 | bolt-arz-DF-175-182188-1048449.arz.su - bolt-arz-DF-221-194675-6456787.arz.su |
| **ARZ8 TEST** | 9 | 7074 | bolt-arz-DF-221-194675-6456952.arz.su - bolt-arz-DF-222-194704-7228783.arz.su |

Table 7: Egyptian Arabic Treebank Data Divisions (ARZ 7-8). All above treebanks are currently version 2.0.

| Division | # Docs | # Words | Document Range |
|---|---|---|---|
| **LEV DEV** | 3 | 3535 | fsa_16902 - fsa_16920 |
| **LEV TRAIN** | 18 | 20871 | fsa_16921 - fsa_17781 |
| **LEV TEST** | 3 | 3030 | fsa_17920 - fsa_18520 |

Table 8: Spoken Levantine ATB Version 2.0 Data Divisions

| Division | # Docs | # Words | Document Range |
|----------|--------|---------|----------------|
| **ATB 3 TRAIN** | 509 | 288046 | ANN20020115.0001 - ANN20021015.0100 |
| **ATB 3 DEVTEST** | 90 | 51664 | ANN20021015.0101 - ANN20021215.0045 |

Table 9: ATB3 Zitouni Data Division

# 5 Previous Data Divisions

This section lists some of the more commonly used, previously defined data divisions used for the MSA Treebanks ATB1, ATB2 and/or ATB3. They are described here for reference.

## 5.1 Zitouni

This data division was defined for ATB3 by Zitouni et al. [10]. This division divides the ATB3 corpora on the document level. The documents are sorted by file name alphabetically; the first 509 documents (about 85% of the words) are set as the TRAIN set, and the remainder is used as a single development-test set (DEVTEST). The Zitouni division is shown in Table 5.1.

It should be noted that the Zitouni TRAIN section starts at the beginning of the sorted document list and is followed by its DEVTEST section. This means that Zitouni TRAIN only has partial overlap with the ATB3 TRAIN division defined in this paper (hereafter referred to as 10-80-10 ATB3). The Zitouni DEVTEST division, while largely overlapping with the 10-80-10 ATB3 TEST defined here, contains 29 files assigned to 10-80-10 ATB3 TRAIN (documents ANN20021015.0101 - ANN20021115.0033).

## 5.2 MADA

This data division was defined for use in training and testing the models used within the MADA (Morphological Arabic Disambiguation and Analysis) tool developed by Habash and Rambow [5]. The splits used in the first MADA paper [4] were different. Initially the MADA division used only a copy of the Zitouni division of ATB3 that had its DEVTEST division divided evenly into DEV and TEST subsets. However, later versions of MADA included the entirety of ATB1 and ATB2 as additional training data (that is, the ATB1 and ATB2 treebanks were not divided). The individual ATB MADA divisions are shown in Table 10.

| Division | # Docs | # Words | Document Range |
|---|---|---|---|
| **ATB1 TRAIN** | 734 | 145386 | 20000715_AFP_ARB.0001 - 20001115_AFP_ARB.0236 |
| **ATB2 TRAIN** | 501 | 144199 | UMAAH_UM.ARB_20010721-e.0018 - UMAAH_UM.ARB_backissue_40-e.0025 |
| **ATB3 TRAIN** | 509 | 288046 | ANN20020115.0001 - ANN20021015.0100 |
| **ATB3 DEV** | 45 | 26359 | ANN20021015.0101 - ANN20021115.0066 |
| **ATB3 TEST** | 45 | 25305 | ANN20021115.0068 - ANN20021215.0045 |

Table 10: ATB1, ATB2, ATB3 MADA Data Division

The MADA ATB3 TRAIN division has the same issue as the Zitouni TRAIN division for 10-80-10 ATB3, in that it starts including documents from the top of the sorted document list, leading to partial TRAIN overlap. However, the MADA ATB3 TEST portion is smaller (by 16 documents) than the 10-80-10 ATB3 TEST portion defined here. This means any system trained on 10-80-10 ATB3 TRAIN can evaluate on MADA ATB3 TEST without concern for evaluating on data seen in training. This allows for meaningful comparisons to previous work that used MADA ATB3 TEST.

## 5.3  JHU - Stanford

This division was defined in the John Hopkins 2005 Workshop [1] and was subsequently picked up by the Stanford Natural Language Processing Group [3]. This split was based on an earlier version from Diab et al. [2]. This division was applied the the MSA ATB1, ATB2, and ATB3 treebanks as a whole. The division sorted the documents in each ATB by document name before combining the treebanks. Then a set of documents from the end of each list was set aside and combined into a single TEST set. Then several documents were selected from the remainder (randomly), and combined into a DEVTEST set. The remaining documents were combined into a single TRAIN set.

Since the documents in the DEVTEST set were selected with a random process, the division cannot be represented as a range in a sorted list of documents.

However, the full document listing (too long to reproduce here) can be found on the Stanford NLP group website. [1]

Because of the way in which the JHU DEVTEST divisions were defined randomly, there is only partial overlap between the corresponding 10-80-10 TRAN and DEV divisions. However, the JHU TEST division is a subset of the combined 10-80-10 ATB1, ATB2 and ATB3 TEST divisions. Like with the MADA case, this means that any system trained on 10-80-10 ATB1, ATB2, and/or ATB3 can be evaluated on JHU TEST without a data sharing issue.

# References

[1] David Chiang, Mona Diab, Nizar Habash, Owen Rambow, and Safiullah Shareef. Parsing Arabic Dialects. In *Proceedings of the European Chapter of ACL (EACL)*, 2006.

[2] Mona Diab, Kadri Hacioglu, and Daniel Jurafsky. Automatic Tagging of Arabic Text: From Raw Text to Base Phrase Chunks. In *Proceedings of the 5th Meeting of the North American Chapter of the Association for Computational Linguistics/Human Language Technologies Conference (HLT-NAACL04)*, pages 149–152, Boston, MA, 2004.

[3] Spence Green and Christopher D. Manning. Better Arabic Parsing: Baselines, Evaluations, and Analysis. In *Proceedings of the 23rd International Conference on Computational Linguistics (Coling 2010)*, pages 394–402, Beijing, China, 2010.

[4] Nizar Habash and Owen Rambow. Arabic Tokenization, Part-of-Speech Tagging and Morphological Disambiguation in One Fell Swoop. In *Proceedings of the 43rd Annual Meeting of the ACL*, pages 573–580, Ann Arbor, Michigan, 2005.

[5] Nizar Habash and Owen Rambow. Arabic Diacritization through Full Morphological Tagging. In *Proceedings of the 8th Meeting of the North American Chapter of the Association for Computational Linguistics/Human Language Technologies Conference (HLT-NAACL07)*, 2007.

---

[1] http://nlp.stanford.edu/software/parser-arabic-data-splits.shtml

[6] Mohamed Maamouri, Ann Bies, Tim Buckwalter, Mona Diab, Nizar Habash, Owen Rambow, and Dalila Tabessi. Developing and Using a Pilot Dialectal Arabic Treebank, 2006.

[7] Mohamed Maamouri, Ann Bies, Tim Buckwalter, and Wigdan Mekki. The Penn Arabic Treebank : Building a Large-Scale Annotated Arabic Corpus, 2004.

[8] Mohamed Maamouri, Ann Bies, Sondos Krouna, Fatma Gaddeche, and Basma Bouziri. *Penn Arabic Treebank Guidelines*. Linguistic Data Consortium, 2009.

[9] Mohamed Maamouri, Ann Bies, Seth Kulick, Dalila Tabessi, and Sondos Krouna. Egyptian Arabic Treebank Pilot, 2012.

[10] Imed Zitouni, Jeffrey S. Sorensen, and Ruhi Sarikaya. Maximum Entropy Based Restoration of Arabic Diacritics. In *Proceedings of the 21st International Conference on Computational Linguistics and 44th Annual Meeting of the Association for Computational Linguistics*, pages 577–584, Sydney, Australia, 2006.